\documentclass
[
    a4paper,
    DIV=calc,
    abstract=true,
    11pt,
]
{scrartcl}

\usepackage
{
    amsmath,
    amssymb,
    amsthm,
    authblk,
    dsfont, 
    enumitem,
    graphicx,
    mathtools,
    nicefrac,
    tabularx,
    tikz,
    xcolor,
}
\usepackage[margin=3cm]{geometry}
\definecolor{myGreen}{RGB}{44, 160, 44}
\definecolor{myOrange}{RGB}{255, 127, 14}

\usepackage[scr=boondoxupr]{mathalfa}
\usepackage{algpseudocode}
\usepackage{algorithm}
\usepackage{graphicx}
\usepackage[utf8]{inputenc}

\usepackage[position=top]{subfig}
\usepackage{float}
\usepackage{lipsum}

\numberwithin{equation}{section}

\usepackage[pdffitwindow=false,
            plainpages=false,
            pdfpagelabels=true,
            pdfpagemode=UseOutlines,
            pdfpagelayout=SinglePage,
            bookmarks=false,
            colorlinks=true,
            hyperfootnotes=false,
            linkcolor=blue,
            urlcolor=blue!30!black,
            citecolor=green!50!black]{hyperref}

\usepackage[justification=justified,format=plain]{caption}

\usepackage{color}
\usepackage[labelfont=bf]{caption}
\usetikzlibrary{arrows, arrows.meta, shapes, fit, automata, positioning}

\graphicspath{{pics/}}

\DeclareMathAlphabet{\mathpzc}{OT1}{pzc}{m}{it}

\newcommand{\R}{\mathbb{R}}                                      
\newcommand{\ts}{\hspace*{0.1em}}                                

\newcolumntype{C}[1]{>{\centering\let\newline\\\arraybackslash\hspace{0pt}}m{#1}}

\DeclareMathOperator{\tr}{tr}

\newtheorem{theorem}{Theorem}[section]

\newtheorem{definition}[theorem]{Definition}
\theoremstyle{definition}
\newtheorem{example}[theorem]{Example}

\makeatletter
\renewcommand*\env@matrix[1][*\c@MaxMatrixCols c]{%
  \hskip -\arraycolsep
  \let\@ifnextchar\new@ifnextchar
  \array{#1}}
\makeatother

\mathtoolsset{centercolon} 

\allowdisplaybreaks

\DeclareCaptionLabelFormat{period}{#1~#2}
\makeatletter
\def\blfootnote{\gdef\@thefnmark{}\@footnotetext}
\makeatother
\begin{document}

\title{Optimization of randomized neural networks for transfer operator approximation}

\author[1]{Mohammad Tabish\thanks{Corresponding author: \href{mailto:M.Tabish-1@sms.ed.ac.uk}{M.Tabish-1@sms.ed.ac.uk}}}
\author[2]{Stefan Klus}
\affil[1]{Maxwell Institute for Mathematical Sciences, University of Edinburgh and Heriot--Watt University, Edinburgh, UK}
\affil[2]{School of Mathematical \& Computer Sciences, Heriot--Watt University, Edinburgh, UK}

\date{}
\maketitle

\begin{abstract}
RaNNDy is a randomized neural network architecture for the data-driven approximation of transfer operators associated with complex dynamical systems. The weights and biases of the hidden layers of the network are randomly initialized and kept fixed, only the output layer is trained. This has several advantages over fully optimized neural networks, notably a closed-form solution for the output layer and significantly lower training costs. Despite these advantages, RaNNDy is restricted to the initial selection of weights and biases that parametrize the basis functions required for the operator approximation. Since the basis functions are determined by the activation function, choosing an appropriate activation function for the hidden layers is crucial. In this work, we propose an algorithm that optimizes the activation function itself, while keeping the weights and biases in the randomized neural network fixed, providing a more suitable dictionary. We illustrate the efficacy of the approach using various benchmark problems, including stochastic differential equations and random walks on graphons.
\end{abstract}

\section{Introduction}
\label{sec:introduction}

Data-driven methods for the approximation of transfer operators have become an area of intense research, as they enable the analysis of the global behavior of complex dynamical systems from data. Spectral decompositions of these operators can be used to analyze the long-term behavior of dynamical systems, e.g., metastable states of molecules or coherent structures in fluid flows, see, e.g., \cite{klus2024dynamical} for an overview of different applications. Existing data-driven methods such as \emph{extended dynamic mode decomposition} (EDMD) \cite{williams2015data, KKS16} and the \emph{variational approach of conformation dynamics} (VAC) \cite{noe2013variational, nuske2014variational} learn a finite-dimensional representation of transfer operators using a fixed set of basis functions, also known as a dictionary. The dictionary could, for instance, comprise monomials, trigonometric functions, Gaussian functions, or indicator functions. The optimal choice of basis functions, however, is highly system-dependent and in general an open problem. Neural networks (NNs) have been used to represent the dictionary required to approximate the operators, and several architectures have been proposed in the literature, see, e.g., \cite{li2017extended, enoch2019, mardt2018vampnets}. The training of these networks is performed using backpropagation, which iteratively minimizes a loss function to update the networks' parameters. This approach is computationally expensive for complex networks and can also lead to problems such as slow convergence, numerous local minima, exploding/vanishing gradients, and sensitivity to hyperparameters such as the learning rate~\cite{gori1992problem, TEBRAAKE199571}. Randomized neural networks avoid using backpropagation by fixing the weights and biases of the hidden layers and training only the output layer for a particular task \cite{zhang2016survey, malik2023random}, making the training computationally cheaper. Based on this idea, we recently proposed RaNNDy \cite{tabish2025deep}, a randomized neural network for the data-driven approximation of transfer operators. RaNNDy computes eigenfunctions of the operator directly via a closed-form solution for the output layer.

Fixing the hidden layers' weights and biases, however, leaves no scope for the optimization of the randomized basis. That is, if the randomly generated basis is not suitable, i.e., it is not able to represent the eigenfunctions of the operator accurately, it could lead to poor approximations. Hence, we propose an algorithm to optimize hyperparameters of parametrized activation functions while keeping the parameters of the model fixed. In other words, we optimize the network while maintaining the computational advantages of fixed weights and biases. We use the variational principle to optimize the activation parameters. The main contributions of this work are:
\begin{enumerate}
    \item We propose a novel algorithm to optimize the randomized basis of RaNNDy by tuning the hyperparameters of the activation function, while keeping the network weights and biases fixed.
    \item We present various applications, including the Bickley jet, high-dimensional protein folding processes, and random walks on graphons.
\end{enumerate}
The proposed algorithm can be regarded as a compromise between fully randomized and fully trained neural networks for transfer operator approximation. The paper is structured as follows: In Section \ref{sec:background}, we will introduce transfer operators and randomized neural networks. In Section \ref{sec:optimize_ranndy}, we will discuss the proposed algorithm to optimize the randomized basis used by RaNNDy. Numerical results will be outlined in Section \ref{sec:numerical_experiments} and a conclusion will be presented in Section \ref{sec:conclusion}.

\section{Background}
\label{sec:background}

We start by briefly introducing transfer operators for reversible and non-reversible dynamical systems and randomized neural networks.

\subsection{Transfer operators}

Transfer operators offer an alternative viewpoint for analyzing stochastic processes in terms of observables and probability densities instead of states of the system. The Koopman operator, for example, describes the evolution of observables and the Perron--Frobenius operator the evolution of the probability densities of the system. These operators are infinite-dimensional but linear representations of the underlying finite-dimensional but nonlinear dynamics. Since we cannot work with infinite-dimensional operators in practice, the goal is to find suitable finite-dimensional approximations of these operators.

Let $\{X_t\}$ be a stochastic process defined on the state space $\mathbb{X} \subset \mathbb{R}^d$ that is governed by a \emph{stochastic differential equation} (SDE) of the form
\begin{equation}\label{eq:sde}
    \mathrm{d}X_t = b(X_t) \ts \mathrm{d}t + \sigma(X_t) \ts \mathrm{d}W_t,
\end{equation}
where $ b \colon \mathbb{R}^d \rightarrow \mathbb{R}^d $ is the drift term, $\sigma \colon \mathbb{R}^d \rightarrow \mathbb{R}^{d \times d}$ is the diffusion term, and $ W_t $ is a $ d $-dimensional Wiener process. In what follows, let $p_{\tau}\colon \mathbb{X} \times \mathbb{X} \rightarrow \mathbb{R}$ be the conditional probability of $X_{t+\tau} = y$ given $X_t = x$, where $\tau$ is a fixed lag time. Additionally, let $L^r$ denote the space of (equivalence classes of) $r$-Lebesgue integrable functions and $L^r_\mu$ the corresponding $\mu$-weighted space, with $1 \leq r \leq \infty$.
\begin{definition}[Transfer operators]
    Let $\tau > 0$ be a fixed lag time.
    \begin{enumerate}
        \item The \emph{Perron--Frobenius operator} $\mathcal{P}^\tau$ is defined by
        \begin{align*}
        \mathcal{P}^\tau \rho(x) = \int_{\mathbb{X}}p_{\tau}(y, x) \ts \rho(y) \ts \mathrm{d}y.
        \end{align*}
        \item The \emph{Koopman operator} $\mathcal{K}^\tau$ is defined by
        \begin{align*}
        \mathcal{K}^\tau f(x) = \int_{\mathbb{X}}p_{\tau}(x, y)f(y) \ts \mathrm{d}y.
        \end{align*}
        \item The reweighted \emph{Perron--Frobenius operator} $\mathcal{T}^\tau$  that propagates densities w.r.t.\ a reference density $\mu$ is defined by
    \begin{align*}
        \mathcal{T}^{\ts\tau} \ts u(x) = \frac{1}{\nu(x)} \int_{\mathbb{X}} p_{\tau}(y, x) u(y) \mu(y) \ts \mathrm{d}y,
    \end{align*}
    where $\nu = \mathcal{P}^\tau\mu$.
    \end{enumerate}
\end{definition}
It is well known that the eigenvalues $ \lambda_i $ and the corresponding eigenfunctions $ \varphi_i $ of these transfer operators---or compositions thereof---can help us identify, for example, metastable states in molecular dynamics or coherent sets in fluid flows \cite{mezic2005spectral, schutte2013metastability, Froyland13, KKS16, banisch2017understanding}.

\begin{definition}[Invariant density]
The eigenfunction of $\mathcal{P}^\tau$ corresponding to the eigenvalue $\lambda = 1$ is called \emph{stationary density} or \emph{invariant density}, usually denoted by $\pi$, i.e., it satisfies $\mathcal{P}^\tau \pi = \pi$.
\end{definition}

The operators introduced above are well-defined as $\mathcal{P}^\tau\colon L^2_{1/\pi} \rightarrow L^2_{1/\pi}$, $\mathcal{T}^\tau \colon L^2_\pi \rightarrow L^2_\pi$ and $\mathcal{K}^\tau\colon L^2_{\pi} \rightarrow L^2_{\pi}$, if $\pi$ exists. We refer the reader to \cite{koltai2018optimal} for more details on the domains.

\begin{example}
For the overdamped Langevin equation given by
\begin{align*}
  \mathrm{d}X_t = -\nabla V(x) \ts \mathrm{d}t + \sqrt{2 \beta^{-1}} \ts \mathrm{d}W_t,
\end{align*}
where $V(x)$ is a confining potential and $\beta > 0$ is the inverse temperature, there exists a unique invariant density given by $\pi(x) = e^{-\beta V(x)}$.
\end{example}

\begin{definition}[Reversibility]
The process $X_t$ is called reversible if the so-called \emph{detailed balance condition} is satisfied, i.e.,  $\pi(x) \ts p_\tau (x,y) = \pi(y) \ts p_\tau(y, x) $ for all $x, y \in \mathbb{X}$.
\end{definition}

For reversible processes, the Koopman operator $\mathcal{K}^\tau$ and the Perron--Frobenius operator $\mathcal{P}^\tau$ are self-adjoint with respect to suitably reweighted inner products \cite{nuske2014variational}. However, for non-reversible dynamical systems, they are in general not self-adjoint. That is, their eigenvalues and eigenfunctions may be complex-valued. In this case, we approximate the singular values and singular functions of these operators to analyze the system. The dominant singular functions of these operators allow us to detect coherent sets in fluid flows~\cite{banisch2017understanding}.

\subsection{Randomized neural networks for transfer operators}

In randomized neural networks, the weights and biases of the hidden layers are randomly initialized and kept fixed, only the output layer is trained. The advantage is that the network can be trained without using backpropagation \cite{malik2023random}. The hidden layers act as a \emph{random feature map} (RFM) through which the input data are transformed into a random feature space, and the output is represented using the output layer. For a single hidden layer and an output layer, the model can be understood as a function $\textbf{f}: \mathbb{R}^d \rightarrow \mathbb{R}^n$ that produces $n$ outputs for a $d$-dimensional input data point and is defined by
\begin{align*}
    \textbf{f}(x) = W_o \cdot \sigma(W x + b),
\end{align*}
where $\sigma$ is the activation function applied component-wise to the vector, $W \in \mathbb{R}^{N \times d}$ are the weights of the hidden layer for $N$ neurons, $b \in \mathbb{R}^N$ is the bias term, and $W_o \in \mathbb{R}^{n \times N}$ is the matrix containing the output weights for the $n$ outputs. Training the network now involves finding the optimal output layer weights $W_o$ \cite{malik2023random, zhang2016survey}.

\section{Optimizing the activation function in RaNNDy}
\label{sec:optimize_ranndy}

Consider a set of $N$ basis functions $\psi_i \colon \mathbb{X} \rightarrow \mathbb{R} $, with $i = 1,\dots, N$, written as a vector-valued function
\begin{equation*}
    \psi(x) = [\psi_1(x), \psi_2(x), \dots, \psi_N(x)]^\top.
\end{equation*}
Provided that these functions are linearly independent, the basis spans an $N$-dimensional subspace for the finite-dimensional approximation of the Koopman operator $ \mathcal{K}^\tau $ projected onto this subspace, denoted by $\mathcal{K}_\psi^\tau$. RaNNDy utilizes the random feature map (RFM) of randomized neural networks as a basis and approximates the $n$ dominant eigenfunctions of the operator, which are then represented by the output layer. RaNNDy is trained using the variational principle to find the weights of the output layer. For a given set of hidden layer weights and biases, the randomized basis can be written as
\begin{equation*}
    \psi(x) = \sigma(Wx + b).
\end{equation*}
However, as mentioned above, fixing the hidden layers does not leave any scope for the optimization of the basis, i.e., for systems where the eigenfunctions cannot be represented by the fixed basis due to the poor initialization of the hidden layers' weights and biases. We thus want to optimize the basis to obtain more accurate approximations of the eigenfunctions.

We can provide flexibility in the selection of randomized basis functions in RaNNDy by introducing a set of tunable parameters in the selected activation function and the distribution of the weights and biases. These parameters could, for instance, be the scale of the distribution from which the hidden weights are sampled. That is, the randomized basis now depends on a parametric activation function $\sigma$, i.e.,
\begin{equation*}
    \psi(x, \omega) = \sigma(W(\omega_W) x + b(\omega_b), \omega_a),
\end{equation*}
where $\omega = [\omega_a, \omega_W, \omega_b]^\top$ contains the parameters to be optimized, $W(\omega_W) \colon \mathbb{R}^{p_W} \rightarrow \mathbb{R}^{N \times d}$ and $b(\omega_b) \colon \mathbb{R}^{p_b} \rightarrow \mathbb{R}^{N}$ are the functions that depend on the parameters $\omega_W$ and $\omega_b$, respectively. For example, $\omega_W$ could be the scale of the standard normal distribution, etc., and $\omega_a \in \mathbb{R}^{p_a}$ depends on the selected activation. For instance, a parametric version of the $\tanh$ activation function would be
$ \psi(x, \omega) = \tanh(W(\omega_W) x + b(\omega_b)) $.

\subsection*{Self-adjoint Koopman operators}

To find the best hyperparameters $\omega$ for the randomized basis $\psi(x, \omega)$, we can define a loss function using the Rayleigh variational principle, proposed in \cite{noe2013variational}, that provides a principled way to approximate spectral decompositions of transfer operators.
\begin{theorem}\label{theorem:variational_principle}
    Let $\widehat{\varphi}_i$ be an approximation of the $i$th true eigenfunction $\varphi_i$ of the Koopman operator $\mathcal{K}^\tau$. Assuming that $\widehat{\varphi}_i$ is normalized and orthogonal to the previous $i - 1$ eigenfunctions, we have
    \begin{align*}
        \langle \mathcal{K}^\tau \widehat{\varphi}_i, \widehat{\varphi}_i \rangle \leq \lambda_i.
    \end{align*}
\end{theorem}
Hence, at each training epoch, for a given basis parametrized by $\omega$, we approximate the eigenvalues of the projected operator and, using the above theorem, maximize their sum to find the best approximation of the eigenfunctions. For a function $f(x) = \sum_{i=1}^n w_i \psi_i(x, \omega)$, we can optimize the Rayleigh quotient
\begin{align*}
    \max_f \frac{\langle f, \mathcal{K}_\psi^\tau f \rangle}{\langle f, f \rangle} = \underset{\substack{w}}{\max} \frac{w^\top C_{01}(\omega) w}{w^\top C_{00}(\omega) w},
\end{align*}
where
\begin{equation*}
     [C_{01}(\omega)]_{ij} = \langle \psi_i (\ts \cdot \ts, \omega), \mathcal{K}_\psi^\tau \psi_j (\ts \cdot \ts, \omega) \rangle, \quad
     [C_{00}(\omega)]_{ij} = \langle \psi_i(\ts \cdot \ts, \omega), \psi_j(\ts \cdot \ts, \omega) \rangle.
\end{equation*}
The optimal solutions $w$ of the above problem are given by the dominant eigenvectors of the matrix representation $A(\omega) = C_{00}^{-1}(\omega) \ts C_{01}(\omega)$ of the projected operator $\mathcal{K}_{\psi}^\tau$, see \cite{noe2013variational, tabish2025deep} for details. These eigenvectors are orthonormal with respect to the inner product weighted by the matrix $C_{00}(\omega)$~\cite{noe2013variational}. Hence, the sum of the eigenvalues, i.e.,
\begin{align*}
    \sum_{i=1}^n \lambda_i =& \max_{\psi_1, \dots, \psi_n} \sum_{i=1}^n \langle \psi_i(x, \omega), \mathcal{K}_\psi^\tau \psi_i(x, \omega) \rangle, \\
    \text{s.t.} \quad & \langle \psi_i(x, \omega), \psi_j(x, \omega) \rangle = \delta_{ij},
\end{align*}
can be optimized to tune the activation parameters $\omega$ using, for example, gradient descent or Bayesian optimization \cite{frazier2018tutorialbayesianoptimization}. After each training step, we get an optimized set of basis function parameters $\omega$. In general, we cannot compute the required integrals in the above equations and thus have to approximate them from data.

For a fixed lag time $\tau$, let $x_i, y_i \in \mathbb{X}$ denote the states of the system, where $x_i = X_t$ and $y_i = X_{t+\tau}$. The data can be obtained by direct measurements or simulations, for example, by numerically solving \eqref{eq:sde}. The obtained data can then be stored in the matrices $X, Y \in \mathbb{R}^{d \times m}$, defined by
\begin{align}
    X = [x_1, x_2, \dots, x_m] \quad \text{and} \quad Y = [y_1, y_2, \dots, y_m].
\end{align}
The matrices $C_{00}(\omega)$ and $C_{01}(\omega)$ can be approximated from data using
\begin{align*}
    \widehat{C}_{00}(\omega) &= \frac{1}{m}\sum_{i=1}^m \psi(x_i, \omega) \ts \psi(x_i, \omega)^\top = \frac{1}{m}\Psi_0(\omega) \ts \Psi_0(\omega)^\top, \\
    \widehat{C}_{01}(\omega) &= \frac{1}{m}\sum_{i=1}^m \psi(x_i, \omega)\psi (y_i, \omega)^\top = \frac{1}{m}\Psi_0(\omega) \ts \Psi_1(\omega)^\top,
\end{align*}
where $\Psi_0(\omega) = \psi(X, \omega) \in \R^{N \times m}$ and $\Psi_1(\omega) = \psi(Y, \omega) \in \R^{N \times m}$, see \cite{tabish2025deep}. Let $\widehat{A}(\omega) = \widehat{C}_{00}(\omega)^{+} \widehat{C}_{01}(\omega)$, finally the loss function to train the network can be written as
\begin{align*}
    \max_{\omega} \tr\left(\widehat{A}(\omega)\right).
\end{align*}
Let $\omega^*$ be the optimal hyperparameters of the basis $\psi(x, \omega)$ after training the hidden layers to construct the covariance and cross-covariance matrices, and let $W_o \in \mathbb{R}^{N \times n}$ be the matrix of the output layer weights for $n$ outputs. To obtain the top $n$ eigenvectors of $\widehat{A}(\omega^*)$ for the approximation of $n$ dominant eigenfunctions of the operator, we can solve the following optimization problem and find a closed-form solution for the output layer
\begin{align*}
    \max_{W_o \in \mathbb{R}^{N \times n} } \quad &
    \text{tr} \left( W_o^\top \widehat{C}_{01}(\omega^*) \ts W_o \right), \\
    \quad \text{s.t.}  \quad &
    W_o^\top \widehat{C}_{00}(\omega^*) W_o = I.
\end{align*}
Using the Lagrange multiplier method and the fact that the Koopman operator is self-adjoint, we get the output layer as the solution of the following eigenvalue problem
\begin{equation*}
    \widehat{C}_{00}^{+}(\omega^*) \ts \widehat{C}_{01}(\omega^*) \ts W_o = W_o \Lambda,
\end{equation*}
see \cite{tabish2025deep} for more details on the solution of the above optimization problem.

\subsection*{Non-self-adjoint Koopman operators}

For non-reversible systems, i.e., those for which the Koopman operator is in general not self-adjoint, RaNNDy approximates the singular values and singular functions of the operator using the variational principle for the singular values \cite{wu2020variational}. Following the above procedure and \cite{tabish2025deep, wu2020variational}, the projected forward--backward operator (a composition of the Koopman operator and its adjoint) is given by $\widehat{A}(\omega) = \widehat{C}_{00}(\omega)^{+} \widehat{C}_{01}(\omega)\widehat{C}_{11}(\omega)^{+}\widehat{C}_{10}(\omega)$ and we obtain the following loss function to tune the activation function of the network and to find the best hyperparameters $\omega$
\begin{align*}
    \max_{\omega} \tr\left(\widehat{A}(\omega)\right).
\end{align*}

Let the columns of $W_o' \in \mathbb{R}^{N \times n}$ and $W_o \in \mathbb{R}^{N \times n}$ denote the top $n$ left and right singular vectors, respectively, that approximate the left and right singular functions of the operator. Then, with the optimal parameters $\omega^*$ of the activation function, RaNNDy solves the following optimization problem to approximate the singular functions corresponding to the largest $ n $ singular values
\begin{align*}
    \max_{W_o, W_o' \in \mathbb{R}^{N \times n}} & \quad \tr (W_o^\top C_{01}(\omega^*) W_o'), \\
    \text{s.t.} & \quad W_o^\top C_{00}(\omega^*) W_o = I, \\
    & \quad W_o'^\top C_{11}(\omega^*) W_o' = I,
\end{align*}
see \cite{tabish2025deep} for a detailed derivation. Again, using the Lagrange multiplier method, we get the following eigenvalue problem to obtain $W_o$
\begin{equation*}
    \widehat{C}_{00}(\omega^{*})^{+} \widehat{C}_{01}(\omega^{*}) \widehat{C}_{11}(\omega^{*})^{+} \widehat{C}_{10} (\omega^{*}) \ts W_o = W_o \Lambda^2.
\end{equation*}
Using $W_o$ and Lagrange optimality conditions, we can find the weights $W_o'$ \cite{tabish2025deep}. The procedure of tuning the activation functions of the randomized neural networks is summarized in Algorithm \ref{alg:ranndy_optimization}.

\begin{algorithm}
\caption{RaNNDy optimization}
\begin{algorithmic}
\State \textbf{Initialization:}
\begin{itemize}
 \item Given the training data $X, Y \in \mathbb{R}^{d \times m}$.
 \item For $k = 0$, parameters $\omega^{(0)}$, learning rate $\eta$, for tuning $\sigma$ (activation function) of the network.
 \item Sample the initial weights and biases of hidden layers from a given distribution with parameters $\omega_W$ and $\omega_b$ to get an initial random basis $\psi(x, \omega)$.
\end{itemize}

\While{not converged}
    \State Construct $\Psi_0(\omega^{(k)})$, $\Psi_1(\omega^{(k)})$.
    \State Compute $\widehat{C}_{00}(\omega^{(k)}), \widehat{C}_{01}(\omega^{(k)}), \widehat{C}_{10}(\omega^{(k)}), \widehat{C}_{11}(\omega^{(k)})$.
    \If{$\mathcal{K}$ is self-adjoint}
        \State $A(\omega^{(k)}) = \widehat{C}_{00}(\omega^{(k)})^{+} \widehat{C}_{01}(\omega^{(k)})$

    \Else
        \State $A(\omega^{(k)}) = \widehat{C}_{00}(\omega^{(k)})^{+} \widehat{C}_{01}(\omega^{(k)})
               \widehat{C}_{11}(\omega^{(k)})^{+} \widehat{C}_{10}(\omega^{(k)})$
    \EndIf
    \State Define $\mathcal{L}(\omega^{(k)}) = \mathrm{tr}(A(\omega^{(k)}))$.
    \State Compute gradient $\nabla_\omega \mathcal{L}(\omega^{(k)})$.
    \State Update $\omega^{(k+1)} = \omega^{(k)} + \eta \nabla_\omega \mathcal{L}(\omega^{(k)})$.
    \State Set $k \leftarrow k + 1$.
\EndWhile

\State $\blacktriangleright$ With optimized $\omega^*$, solve eigenproblem:
\If{$\mathcal{K}$ is self-adjoint}
    \State $\widehat{C}_{00}(\omega^*)^{+} \widehat{C}_{01}(\omega^*) W_o = W_o \Lambda$
\Else
    \State $\widehat{C}_{00}(\omega^*)^{+} \widehat{C}_{01}(\omega^*) \widehat{C}_{11}(\omega^*)^{+} \widehat{C}_{10}(\omega^*) W_o = W_o \Lambda^2$
\EndIf

\State $\blacktriangleright$ Select the top $n$ eigenvectors from the sorted eigenvectors in $W_o$ (according to the eigenvalues) to get the approximation of the $n$ dominant eigenfunctions/singular functions.
\end{algorithmic}
\label{alg:ranndy_optimization}
\end{algorithm}

\section{Numerical Experiments}
\label{sec:numerical_experiments}

We will now show how the proposed approach can be applied to different types of time-series data.

\subsection{Graphons}

\begin{figure}
    \centering
    \subfloat[][\label{subfig:sym_graphon_w}]{\includegraphics[width=.27\textwidth]{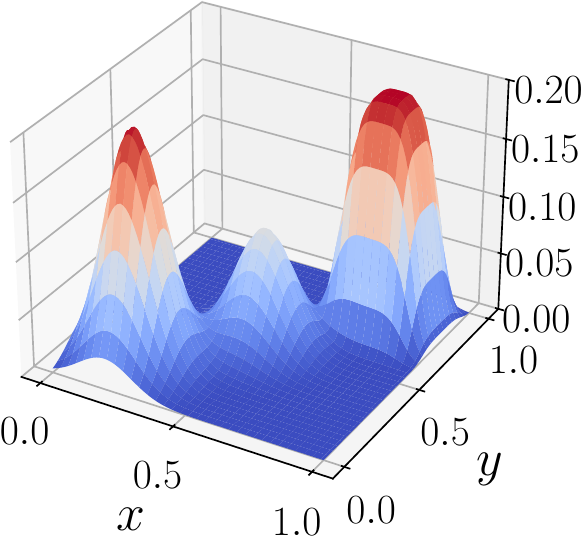}}\quad
    \subfloat[][\label{subfig:sym_graphon_p}]{\includegraphics[width=.27\textwidth]{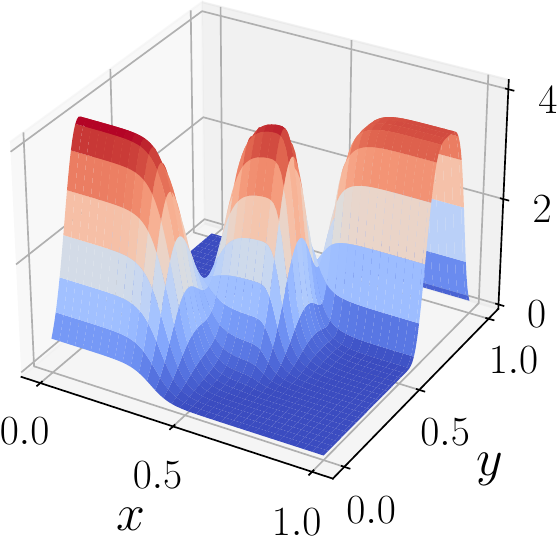}}\quad
    \subfloat[][\label{subfig:sym_graphon_rw}]{\includegraphics[width=.31\textwidth]{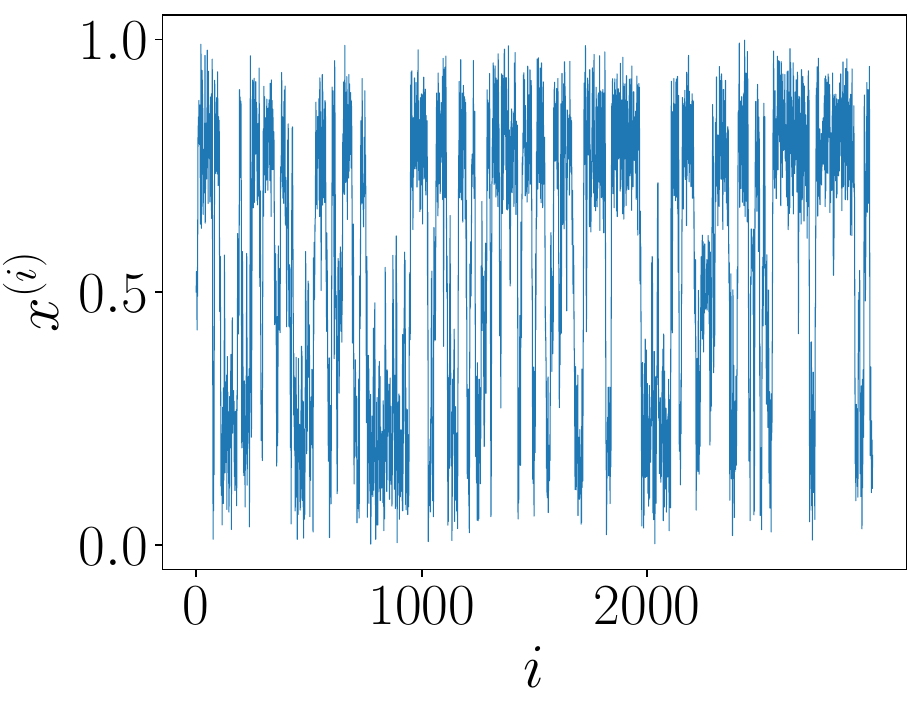}}\quad
    \caption{(a) Symmetric graphon $g$. (b) The associated transition density function $p$. (c)~A random walk on the graphon.}
    \label{fig:sym_graphon}
\end{figure}

\begin{figure}
    \centering
    \subfloat[][\label{subfig:sym_graphon_loss}]{\includegraphics[width=.32\textwidth]{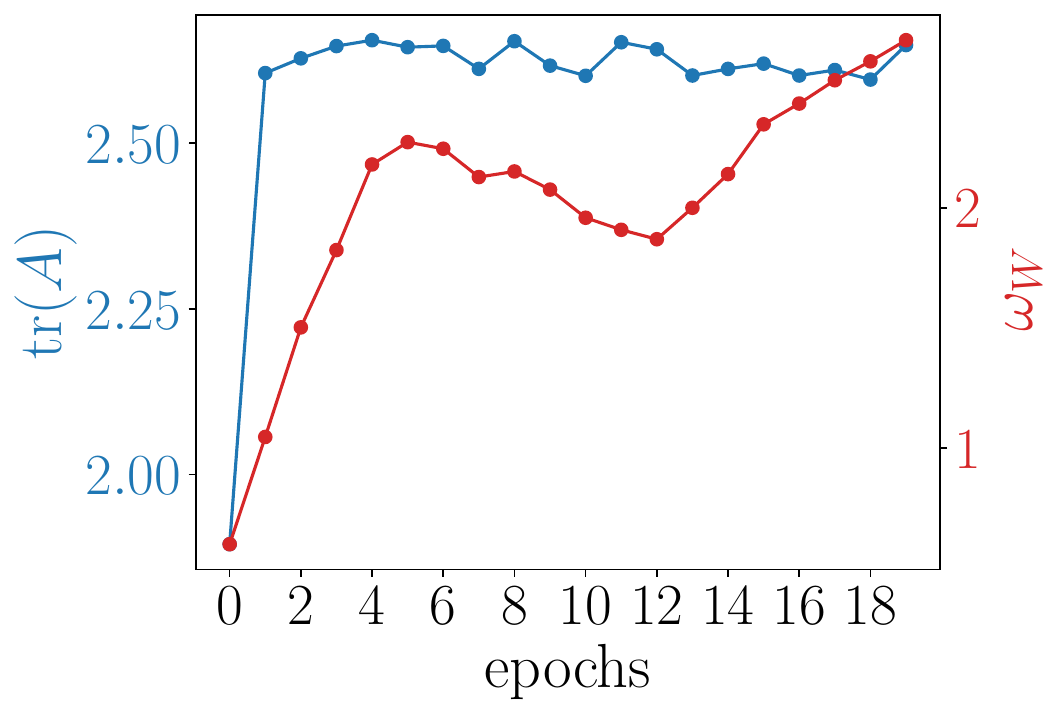}}\quad
    \subfloat[][\label{subfig:sym_graphon_eigvals}]{\includegraphics[width=.31\textwidth]{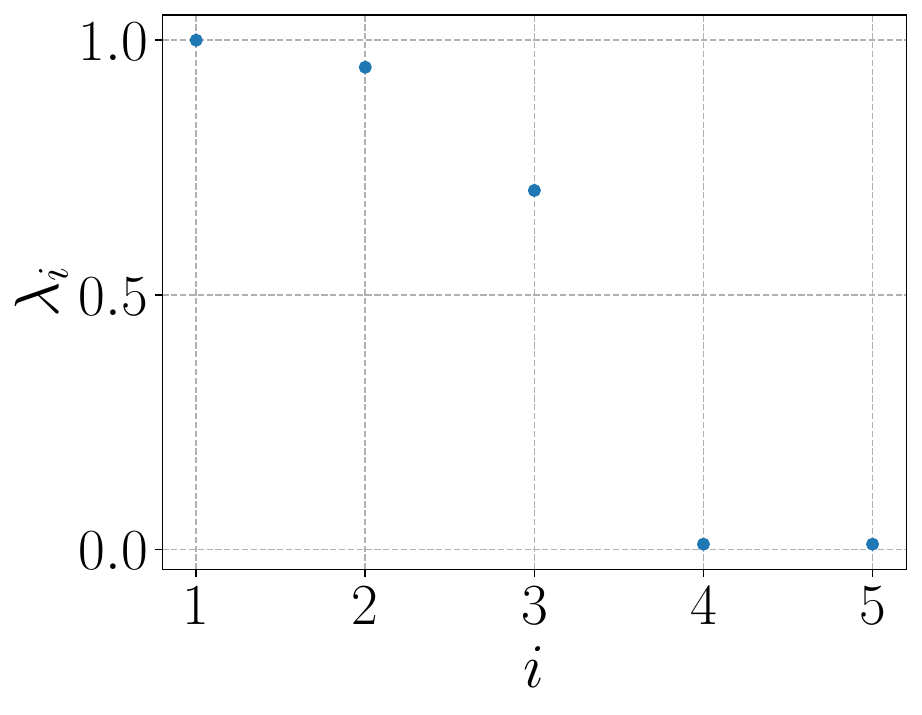}}\quad
    \subfloat[][\label{subfig:sym_graphon_Keigf}]{\includegraphics[width=.31\textwidth]{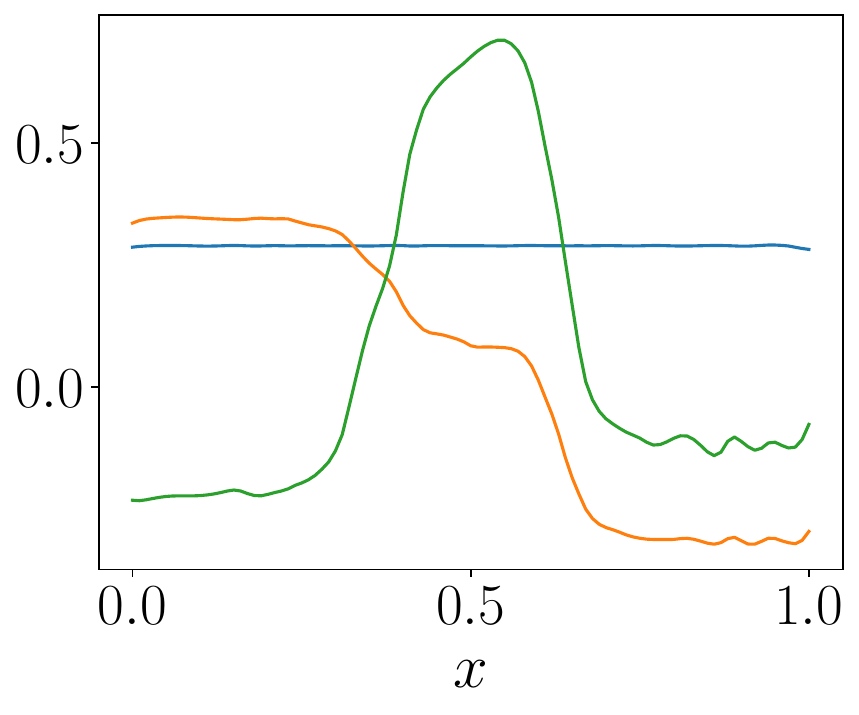}}\quad \\
    \subfloat[][\label{subfig:sym_graphon_Peigf}]{\includegraphics[width=.31\textwidth]{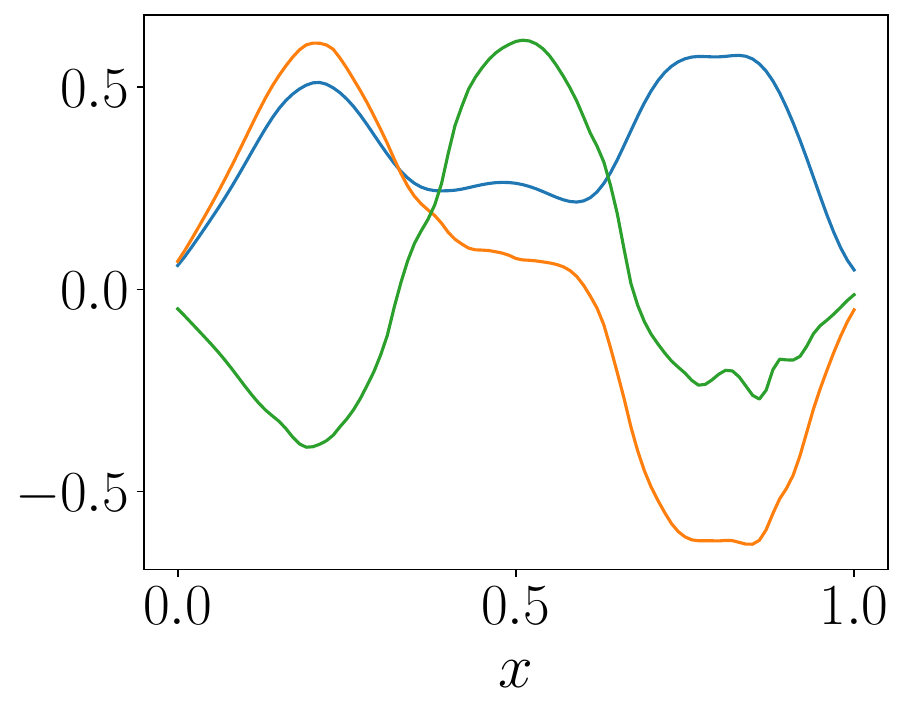}}\quad
    \subfloat[][\label{subfig:sym_graphon_recons_w}]{\includegraphics[width=.29\textwidth]{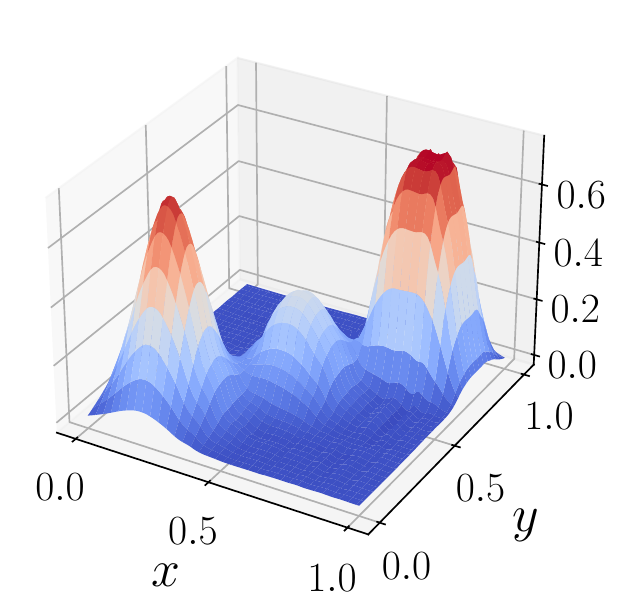}}\quad
    \subfloat[][\label{subfig:sym_graphon_recons_p}]{\includegraphics[width=.29\textwidth]{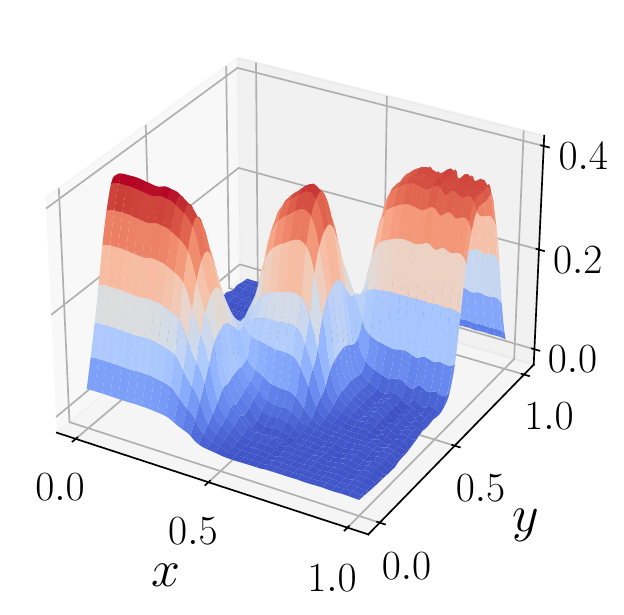}}\quad
    \caption{(a) Training loss for the optimization of RaNNDy with the tanh activation parameter $\omega$. (b) Five dominant eigenvalues of the Koopman operator after training. (c) Three dominant associated Koopman operator eigenfunctions, where \textcolor{blue!30}{\rule[0.5ex]{0.5cm}{1pt}} denotes the first, \textcolor{myOrange}{\rule[0.5ex]{0.5cm}{1pt}} the second, and \textcolor{myGreen}{\rule[0.5ex]{0.5cm}{1pt}} the third eigenfunction. (d) The associated Perron--Frobenius operator eigenfunctions. (e)~Reconstructed graphon $g$ with rank 3. (f) Corresponding transition density $p$.}
    \label{fig:sym_graphon_eigendecomposition}
\end{figure}

Although we only introduced transfer operators for stochastic differential equations, it has recently been shown in \cite{klus2025learning} that these definitions can be easily extended to random walks on graphons. Graphons, which can be interpreted as limits of convergent sequences of graphs, are described by measurable functions $g\colon [0, 1]^2 \rightarrow [0, 1]$, where $g(x, y)$ represents the weight of an edge between the vertices $x, y \in [0, 1]$ (or its probability). A graphon is said to be \emph{symmetric} or \emph{undirected} if $g(x, y) = g(y, x)$. Otherwise, it is called \emph{unsymmetric} or \emph{directed}. Spectral decompositions of transfer operators associated with symmetric graphons are given by
\begin{align*}
    \mathcal{P}\rho = \sum_i \lambda_i \langle\widehat{\varphi}_i, \rho \rangle_{\frac{1}{\pi}} \widehat{\varphi}_i \quad \text{and} \quad \mathcal{K} f = \sum_i  \lambda_i \langle \varphi_i, f \rangle_{\pi} \varphi_i,
\end{align*}
where $\varphi_i$ is an eigenfunction of the Koopman operator and $\widehat{\varphi}_i = \pi \varphi_i$ the corresponding eigenfunction of the Perron--Frobenius operator. This also allows us to reconstruct the transition probability density $p(x, y)$ and the graphon $g(x, y)$, i.e.,
\begin{align*}
    p(x, y) = \sum_i \lambda_i \varphi_i(x) \widehat{\varphi}_i(y) \quad \text{and} \quad g(x, y) = Z \sum_i \lambda_i \widehat{\varphi}_i(x) \widehat{\varphi}_i(y),
\end{align*}
where $\pi(x) = \frac{d(x)}{Z}$, with $Z = \int_0^1 d(x) \mathrm{d}x$, is an invariant density and $d(x)$ is the degree function, see \cite{klus2025learning} for details. Hence, provided that there is a spectral gap after the $i$th eigenvalue such that $ \lambda_{i+1} \approx 0 $ (this is the case if there are $i$ clearly separated clusters), we can approximate the graphon $g$ and the corresponding transition density function $p$ using the dominant eigenfunctions. As a basic example, we consider the symmetric graphon defined in \cite{klus2025learning}, given by
\begin{align*}
    g(x,y) &= 0.2 e^{-\frac{(x-0.2)^2 + (y-0.2)^2}{0.02}} + 0.1 e^{-\frac{(x-0.5)^2 + (y-0.5)^2}{0.02}} + 0.2 e^{-\frac{(x-0.8)^4 + (y-0.8)^4}{0.0005}},
\end{align*}
and visualized in Figure~\ref{fig:sym_graphon} along with the corresponding transition density function and a random walk sampled from the graphon that provides the data for the numerical approximations. To demonstrate the application of the proposed algorithm, we initialize RaNNDy with a parametric tanh activation function. We choose the randomized basis
\begin{align*}
    \psi(x, \omega) = \text{tanh}(W(\omega_W)x + b(\omega_b)),
\end{align*}
where $\omega_W$ and $\omega_b$ are the scales of normal distribution and the parameters to be optimized with the proposed algorithm. Using [256, 512, 256] neurons in the hidden layers and initializing $\omega_W = \omega_b = 0.1$, we use Algorithm \ref{alg:ranndy_optimization} to numerically approximate the top five eigenvalues and eigenfunctions of the associated transfer operators. In Figure \ref{subfig:sym_graphon_loss}, we can see that the initial basis with $\omega = 0.001$ was not suitable and further optimization was required. The algorithm converges within few (less than 20) epochs. As expected, we see a spectral gap after the third eigenvalue in Figure \ref{subfig:sym_graphon_eigvals}, indicating the presence of three ``metastable sets" in the system, which can also be seen from the eigenfunctions of the Koopman operator in Figure \ref{subfig:sym_graphon_Keigf}. Using these three dominant eigenfunctions of the Koopman and the Perron--Frobenius operators, we reconstruct the graphon and the corresponding transition density, shown in Figures \ref{subfig:sym_graphon_recons_w} and \ref{subfig:sym_graphon_recons_p}.

\begin{figure}
    \centering
    \subfloat[][\label{subfig:sym_graphon_gridsearch_dists}]{\includegraphics[width=.32\textwidth]{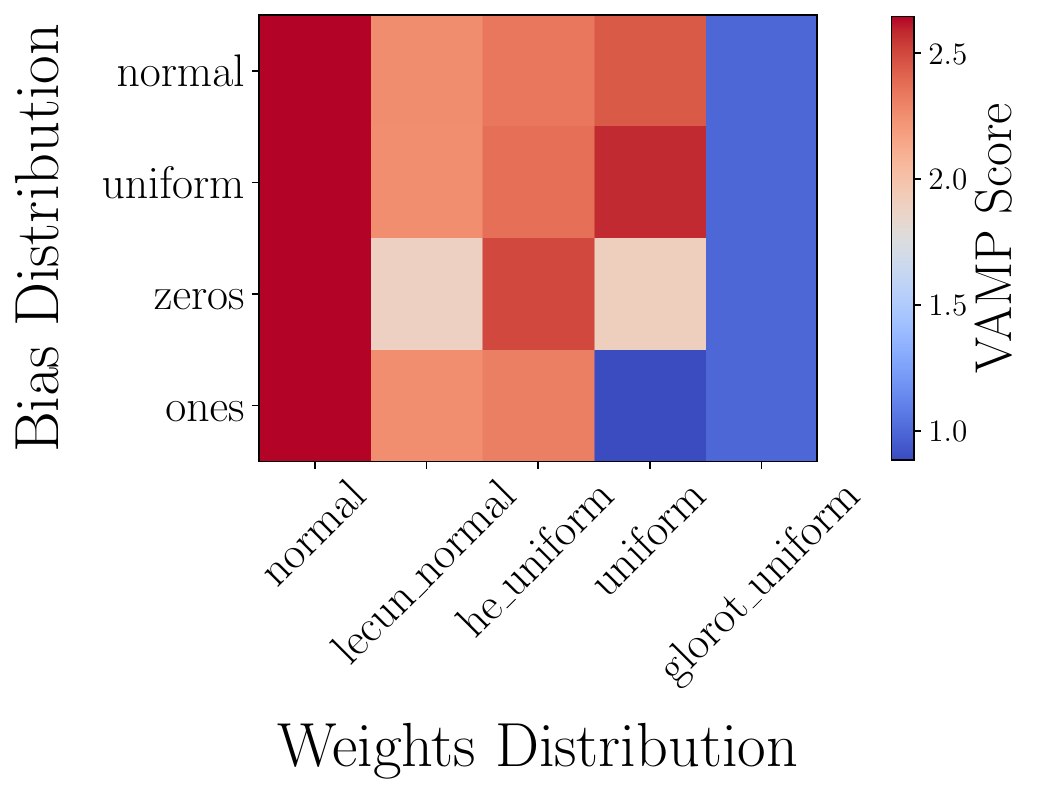}}\quad
    \subfloat[][\label{subfig:sym_graphon_gridsearch_distscales}]{\includegraphics[width=.31\textwidth]{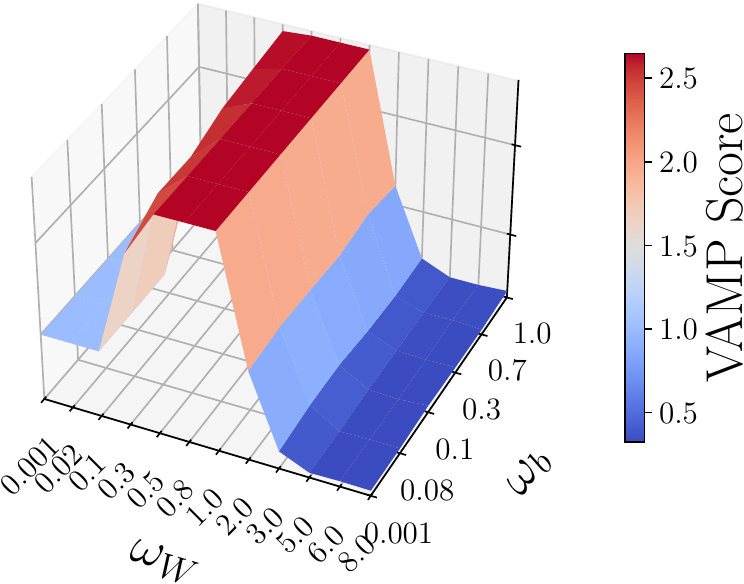}}\quad
     \subfloat[][\label{subfig:ranndy_vampnet_loss_lr1e-2}]{\includegraphics[width=0.31\textwidth]{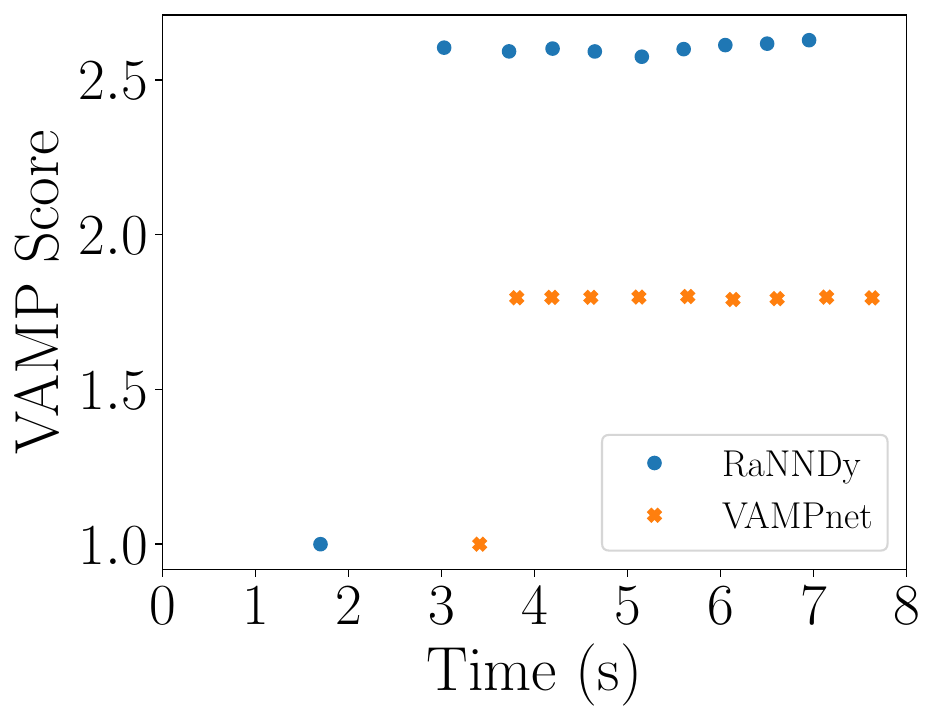}}\quad
    \caption{Selecting the distributions for the hidden layers' weights and biases in RaNNDy. (a) Grid search for different distributions. (b) Loss surface for different values of scales of the normal distributions. (c) Comparison of training the whole network (VAMPnets with learning rate $lr = 10^{-2}$) vs.\ only the scales of the distributions for the initializations of RaNNDy. We can see that the optimizer is stuck in a local minimum, while RaNNDy converges even for a larger learning rate of $lr = 0.5$.}
    \label{fig:sym_graphon_loss_lanscape}
\end{figure}

\paragraph{Comparison:} In Figure~\ref{subfig:sym_graphon_gridsearch_dists}, we show a grid-search method to select the most suitable distributions for the weights and biases for the initialization of RaNNDy, illustrating that the effect of the bias distribution is negligible in this case. In Figure~\ref{subfig:sym_graphon_gridsearch_distscales}, we show the loss surface of the RaNNDy approximation for different values of the scales of the normal distributions for sampling the hidden layers' weights and biases. Figure~\ref{subfig:ranndy_vampnet_loss_lr1e-2} shows the comparison of training the full neural network (VAMPnet) vs.\ tuning the distribution for RaNNDy, showing that RaNNDy converges to a good solution, while the VAMPnet gets stuck in a local minimum. We observed similar behavior for different learning rates and schedulers. For learning rates larger than $10^{-2}$, the VAMPnet approach diverges.

\subsection{Bickley Jet}

\begin{figure}
    \centering
    \subfloat[][\label{subfig:bickley_timestep_0}]{\includegraphics[width=.31\textwidth]{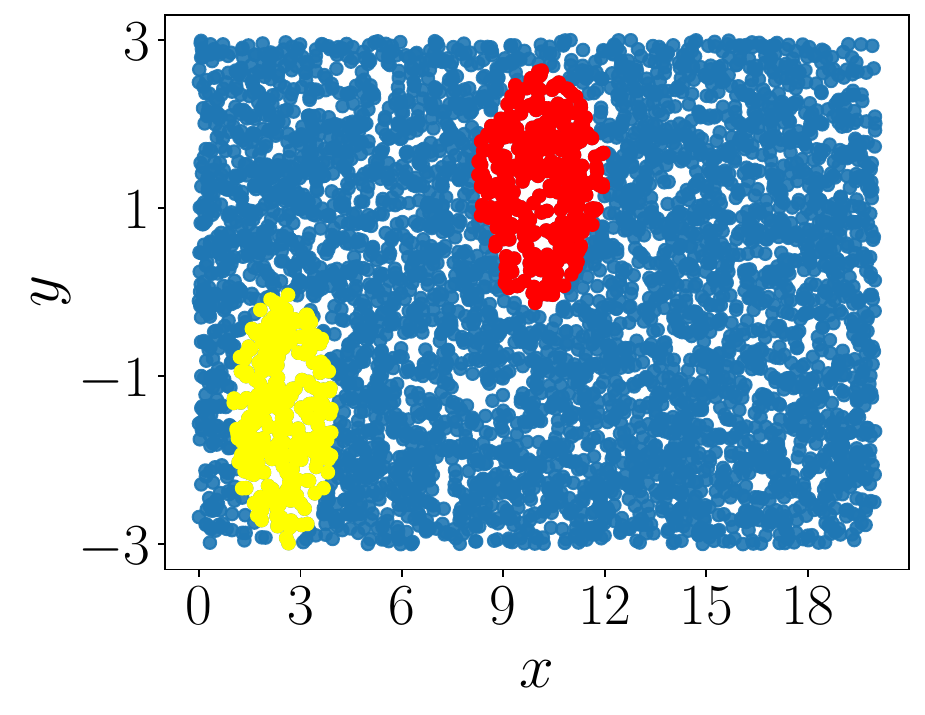}}\quad
    \subfloat[][\label{subfig:bickley_timestep_25}]{\includegraphics[width=.31\textwidth]{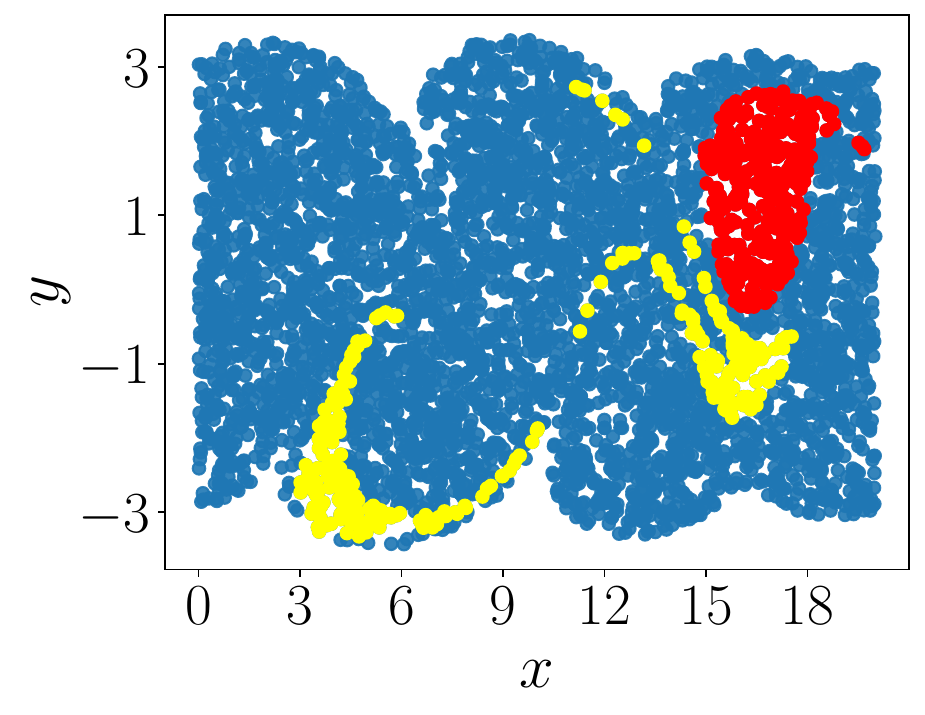}}\quad
    \subfloat[][\label{subfig:bickley_timestep_50}]{\includegraphics[width=.31\textwidth]{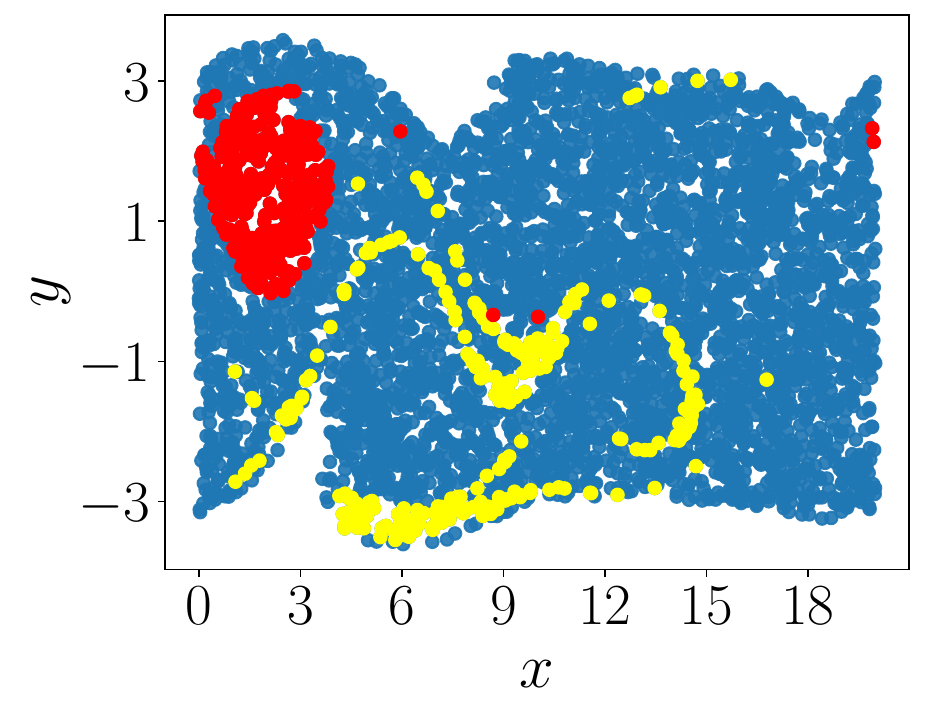}}\quad
    \caption{Bickley jet flow at time (a) $t = 0$, (b) $t = 25$, and (c) $t= 50$.}
    \label{fig:bickley_flow}
\end{figure}

\begin{figure}
    \centering
    \subfloat[][\label{subfig:bickley_loss}]{\includegraphics[width=.45\textwidth]{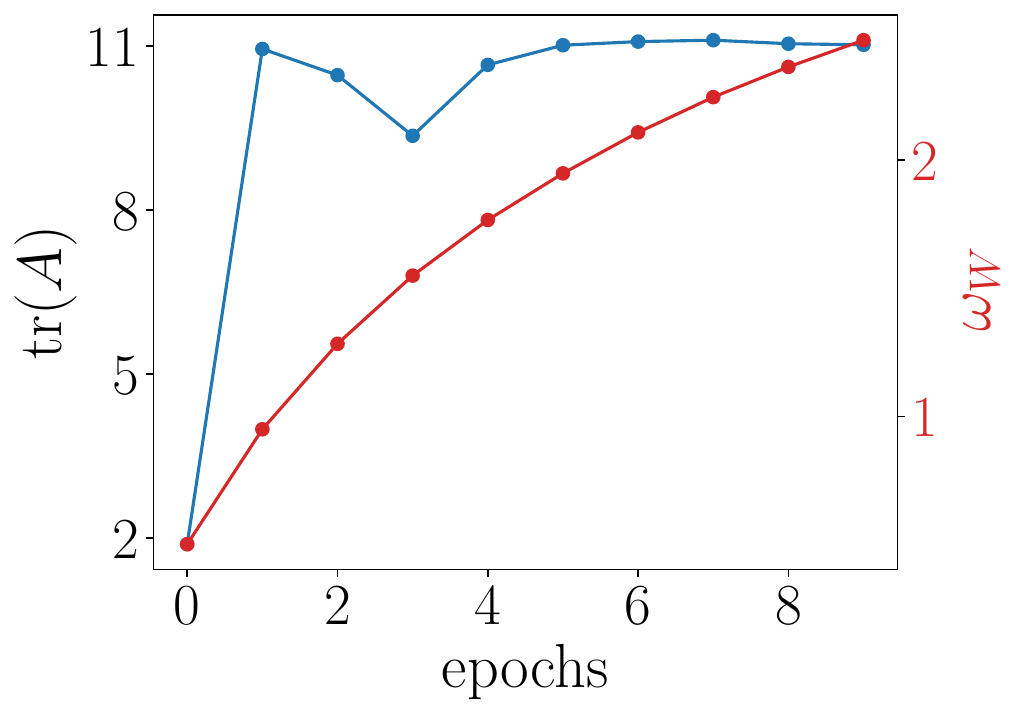}}\quad
    \subfloat[][\label{subfig:bickley_eigvals}]{\includegraphics[width=.42\textwidth]{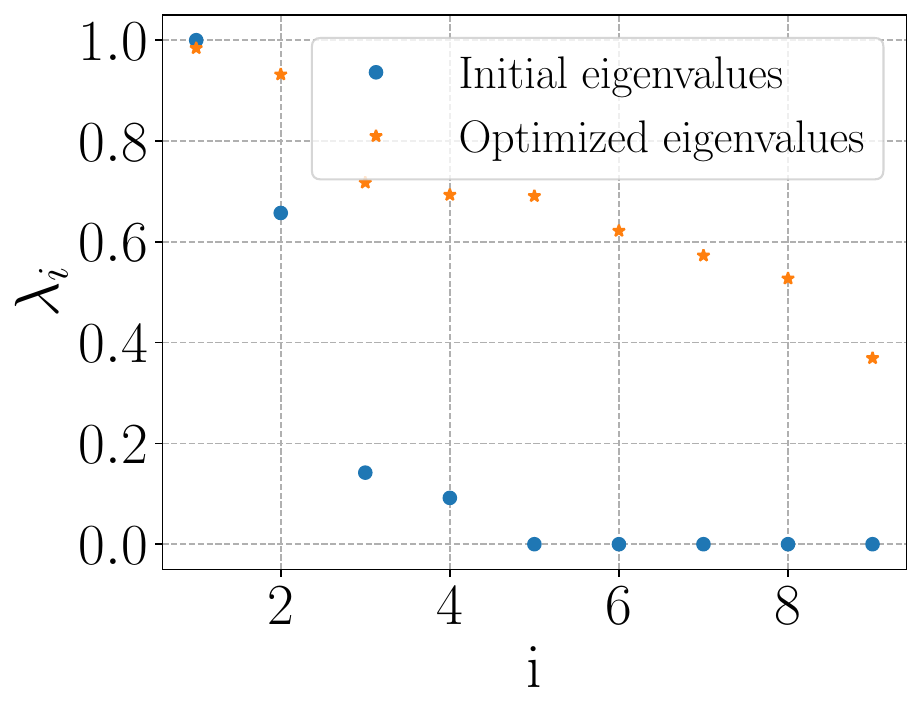}}\quad \\
    \subfloat[][\label{subfig:bickley_initial_clusters}]{\includegraphics[width=.45\textwidth]{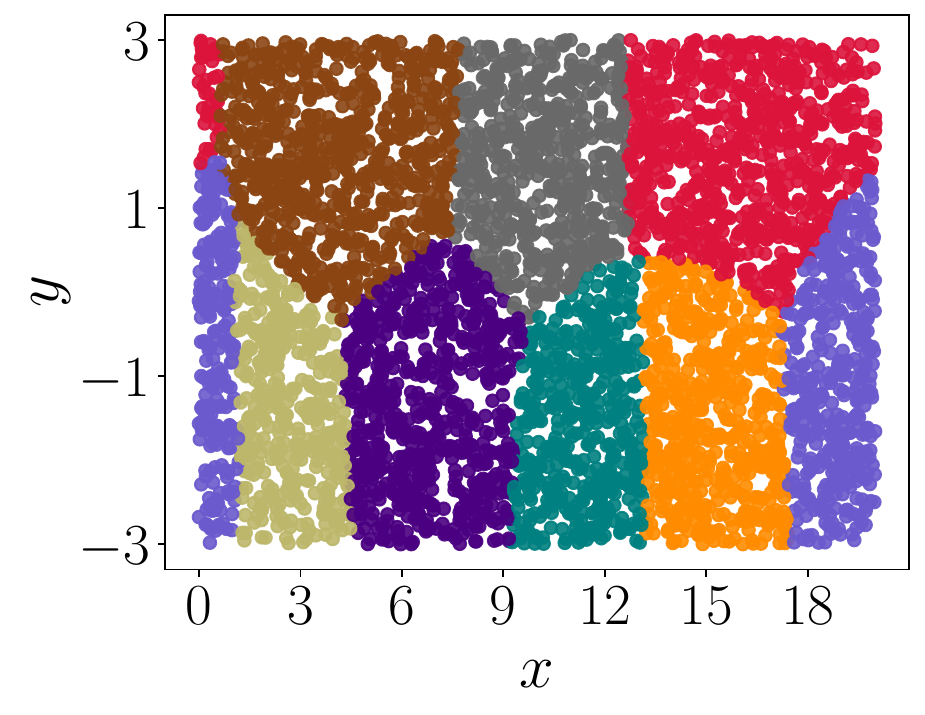}}\quad
    \subfloat[][\label{subfig:bickley_final_clusters}]{\includegraphics[width=.45\textwidth]{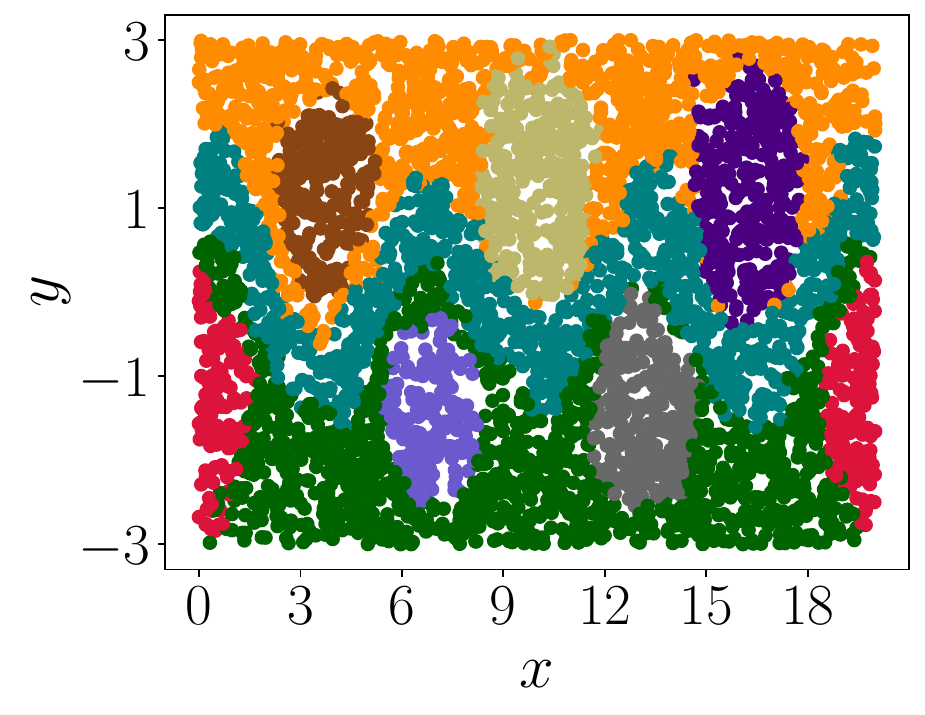}}\quad
    \caption{Optimization of the activation function for the Bickley jet. (a) The loss function and the activation parameter values. (b) Eigenvalues before and after optimization. (c)~Clustering of the initial nine dominant singular functions into nine clusters. (d) Optimized clustering.}
    \label{fig:bickley_results}
\end{figure}

We will now show how we can tune RaNNDy to efficiently detect coherent sets \cite{Froyland13, banisch2017understanding} and apply the proposed algorithm to the Bickley jet, a simple benchmark problem representing an idealized stratospheric flow \cite{rypina2007lagrangian}. The system is given by
\begin{equation*}
    \begin{bmatrix}
    \dot{x} \\
    \dot{y}
    \end{bmatrix}
    =
    \begin{bmatrix}
    - \dfrac{\partial \Phi}{\partial y} \\[2ex]
    \phantom{-} \dfrac{\partial \Phi}{\partial x}
    \end{bmatrix},
\end{equation*}
with the stream function
\begin{align*}
    \Phi(x, y, t) &= c_3 y - U_0 L \tanh\left(\frac{y}{L}\right)
    + A_3 U_0 L \, \text{sech}^2\left(\frac{y}{L}\right) \cos(k_1 x) \\
    & + A_2 U_0 L \, \text{sech}^2\left(\frac{y}{L}\right) \cos(k_2 x - \sigma_2 t) \\
    & + A_1 U_0 L \, \text{sech}^2\left(\frac{y}{L}\right) \cos(k_1 x - \sigma_1 t).
\end{align*}
We generate $m=5 \ts 000$ uniformly sampled initial conditions within the domain $ [0, 20] \times [-4, 4] $. Then, using the methods and parameters from the Python library \emph{deeptime} \cite{hoffmann2021deeptime}, we simulate the flow from $t_0 = 0$ to $t_1 = 40$. A few snapshots of the flow at different times $ t $ are visualized in Figure \ref{fig:bickley_flow}. The particles in red stay close to each other, forming a coherent set, whereas the yellow particles are dispersed by the flow. We again select the $ \tanh $ activation function with parameter $\omega$. The randomized basis is defined by
\begin{align*}
    \psi(x, \omega) = \text{tanh}(W(\omega_W) x + b(\omega_b)).
\end{align*}
Initializing $\omega_W = 0.001$ and $\omega_b = 0.001$ as the scales of the normal distribution from which the weights are sampled, we apply the proposed algorithm to optimize the basis and approximate the dominant nine right singular functions of the operator $ \mathcal{T}^\tau $. The results are presented in Figure \ref{fig:bickley_results}. The algorithm converges in fewer than 10 iterations. The clustering of optimized singular functions into 9 clusters reveals the coherent sets in the Bickley jet.

\subsection{Protein NuG2}

As a last example, we consider high-dimensional molecular dynamics data, where we aim to understand the folding--unfolding dynamics of a protein, see~\cite{Schuette_Klus_Hartmann_2023} for more details. We use simulation data of the protein NuG2 acquired from \href{https://www.deshawresearch.com/}{\color{blue}{D.E. Shaw Research}} \cite{lindorff2011fast}. Protein G (NuG2) is a $56$ residue molecule. The trajectory data has a length of $3.680002 \cdot 10^8$\ts ps. We subsample the trajectory of $929251$ frames to create the training data $\{x_i, y_i\}_{i=1}^m$. With a lag time of $50$ frames between each $x_i$ and $y_i$, we compute the contact maps (distances between the residue pairs of the molecule) and store them in the data matrices $X, Y \in \R^{1431 \times 18 \ts 585}$, here we ignore the near-neighbor residues and self-contact distances.

\begin{figure}
    \centering
    \begin{minipage}[t]{0.45\textwidth}
        \subfloat[][\label{subfig:nug2_eigvals}]{\includegraphics[width=0.8\textwidth]{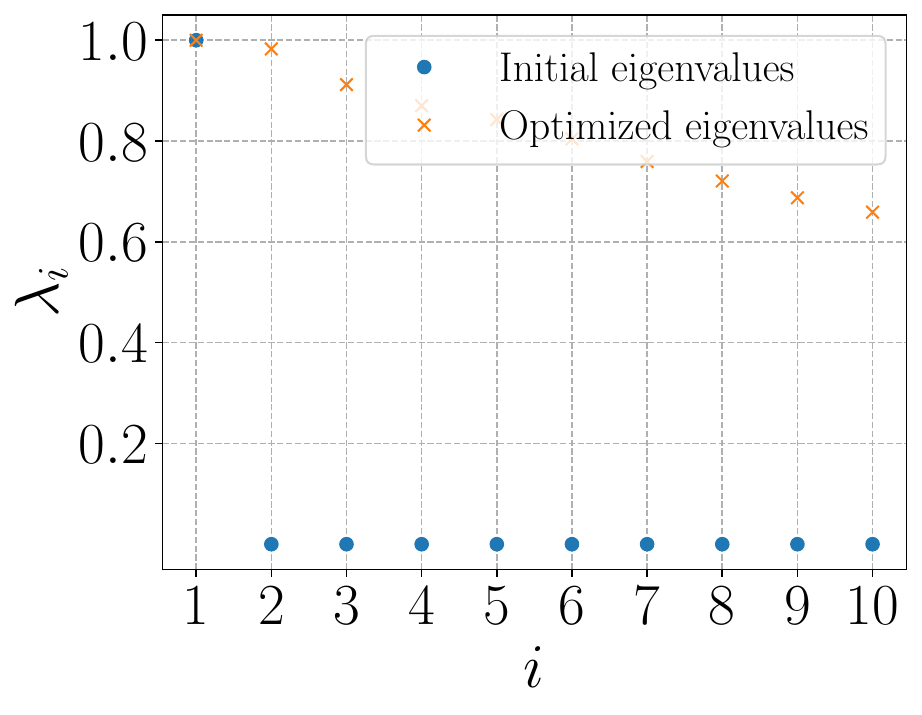}
    }
    \end{minipage}
    \begin{minipage}[t]{0.45\textwidth}
        \subfloat[][\label{subfig:nug2_eigfuncs}]{\includegraphics[width=0.8\textwidth]{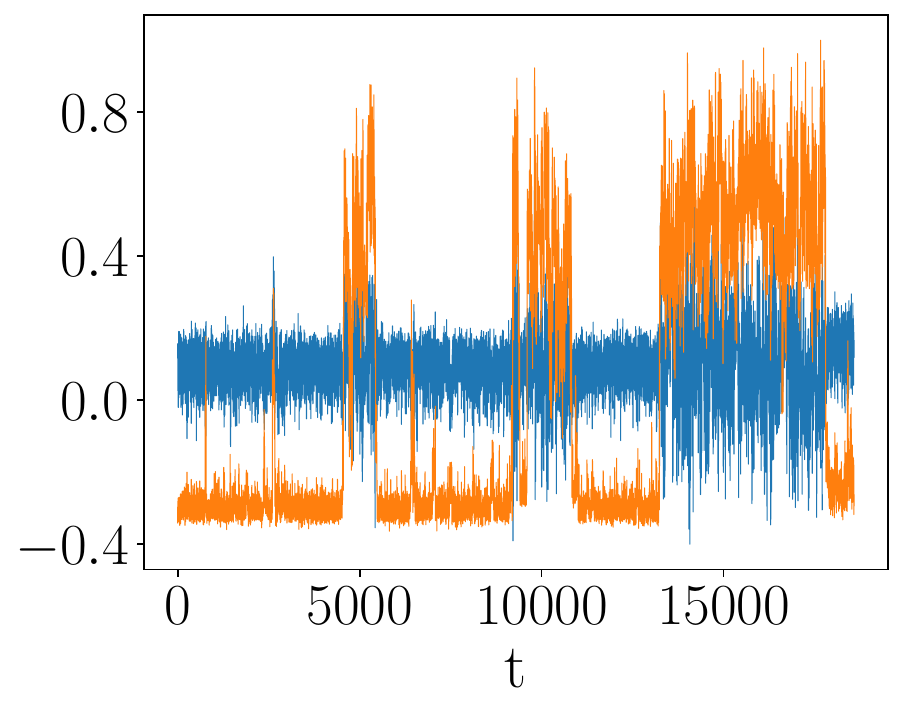}}
    \end{minipage} \\
    \begin{minipage}[t]{0.40\textwidth}
        \subfloat[][\label{subfig:nug2_folded_states}]{\includegraphics[width=0.8\textwidth]{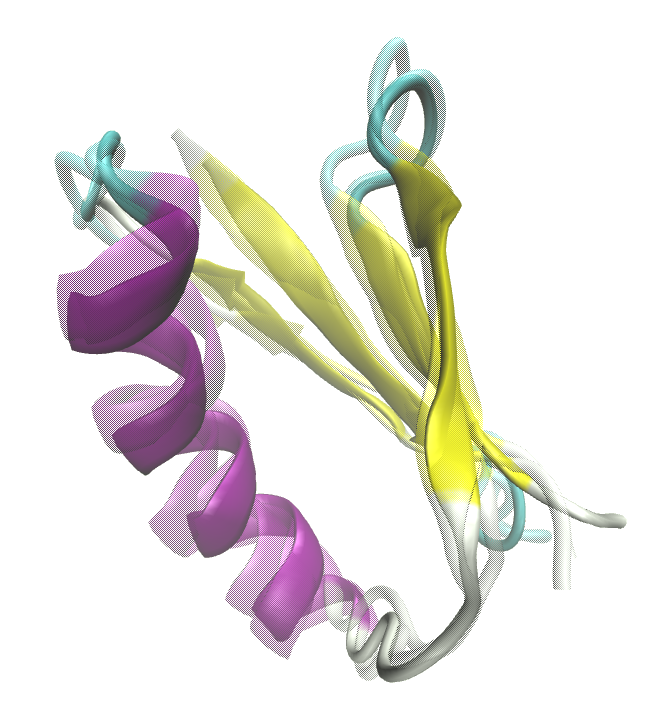}
    }
    \end{minipage}
    \begin{minipage}[t]{0.40\textwidth}
        \subfloat[][\label{subfig:nug2_unfolded_states}]{\includegraphics[width=0.8\textwidth]{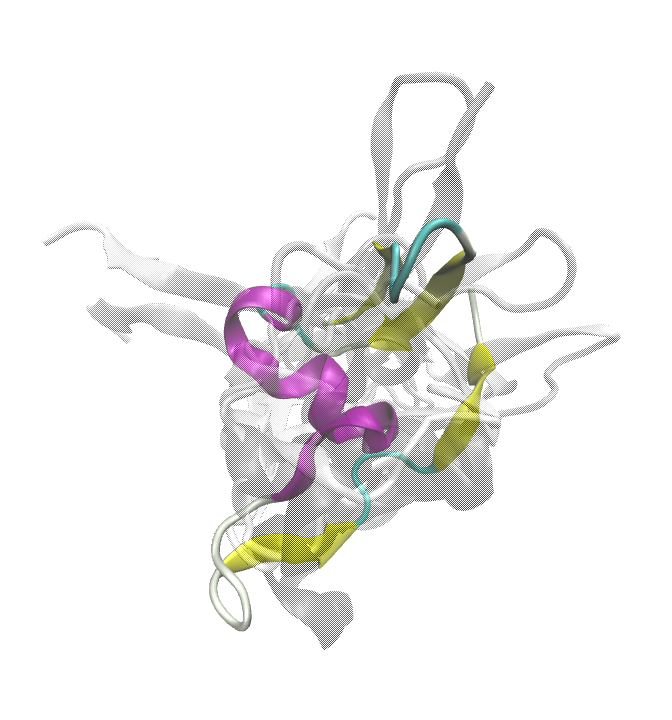}}
    \end{minipage} \\
    \begin{minipage}[t]{0.45\textwidth}
        \subfloat[][\label{subfig:nug2_folded_contact}]{\includegraphics[width=0.9\textwidth]{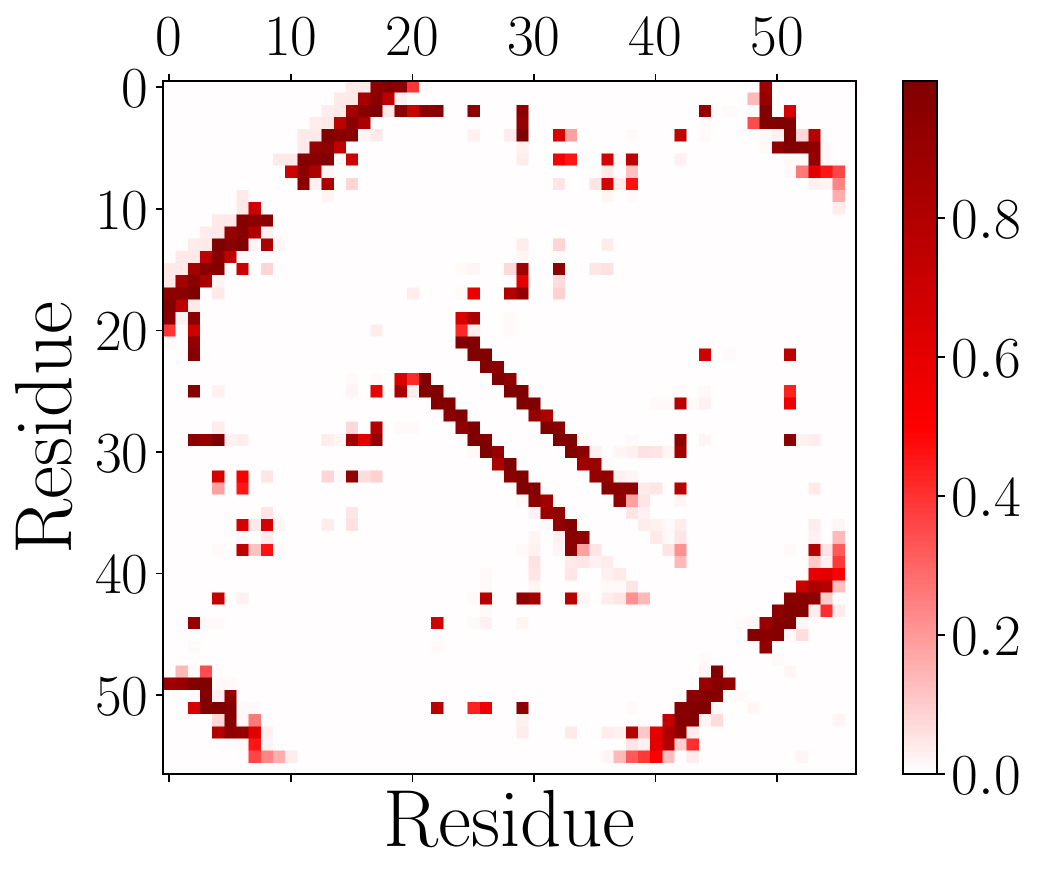}}
    \end{minipage}
    \begin{minipage}[t]{0.45\textwidth}
        \subfloat[][\label{subfig:nug2_unfolded_contact}]{\includegraphics[width=0.9\textwidth]{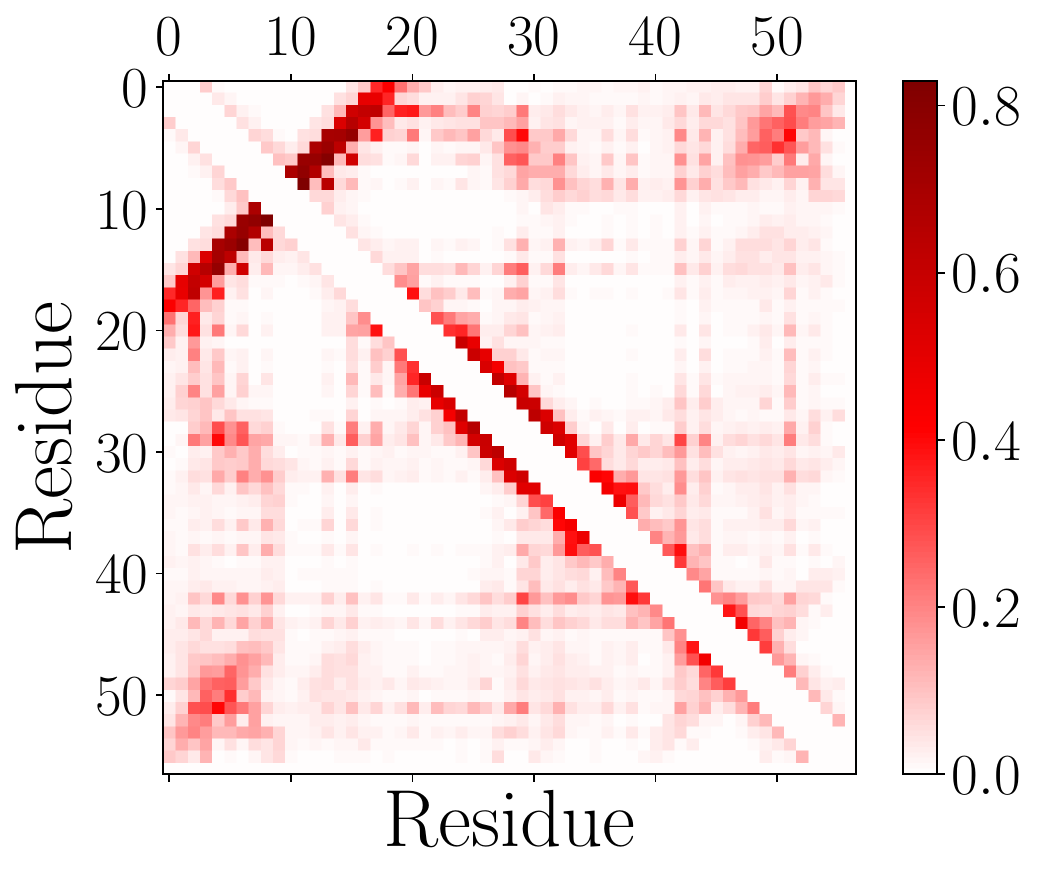}}
    \end{minipage}
    \caption{(a) Ten dominant eigenvalues of the approximated Koopman operator. (b)~The second dominant eigenfunction to distinguish the folded and unfolded states, where \textcolor{blue!90}{\rule[0.5ex]{0.5cm}{1pt}} denotes the initial and \textcolor{myOrange}{\rule[0.5ex]{0.5cm}{1pt}} denotes the optimized eigenfunction evaluated at the data points. (c) \& (d) Some folded and unfolded states of the NuG2 protein extracted using the second eigenfunction. (e) \& (f) Contact map frequencies between different residue pairs over all the identified folded and unfolded states, respectively.}
    \label{fig:nug2_results}
\end{figure}

To approximate the eigenvalues and eigenfunctions of the Koopman operator, we optimize the randomized basis of RaNNDy using the proposed algorithm. With the optimized basis, we approximate the dominant eigenvalues of the Koopman operator, see Figure \ref{subfig:nug2_eigvals}. The second dominant eigenfunction in Figure \ref{subfig:nug2_eigfuncs} represents the two states of the molecule. We can see that the initial randomized basis leads to poor approximation results. In Figures \ref{subfig:nug2_folded_states} and \ref{subfig:nug2_unfolded_states}, we show a few folded and unfolded states along with the contact frequencies in Figures \ref{subfig:nug2_folded_contact} and \ref{subfig:nug2_unfolded_contact} of the NuG2 protein molecule identified using optimized RaNNDy. The contact map shows the stability of the identified folded and unfolded states.

\section{Conclusion and future work}
\label{sec:conclusion}

We proposed an algorithm to optimize RaNNDy, a randomized neural network framework for transfer operator approximation. RaNNDy allows for a closed-form solution for the output layer of the network representing the eigenfunctions of the transfer operator, making it computationally cheap while mitigating common issues associated with deep learning, such as slow convergence. We further improved the benefits of RaNNDy by making it optimizable, i.e., having parametric activation functions that can be tuned. Instead of training all the weights and biases of the hidden layers of the network, we introduce a few hyperparameters in the activation function and tune them to find suitable weights. This can be regarded as a compromise between fully trained NNs such as VAMPnets and RaNNDy, which is fully randomized. We illustrated the efficiency and accuracy of the proposed approach with the aid of several numerical examples, showing that the optimization of the randomized basis functions is important. Furthermore, tuning the distributions of the hidden weights and biases is easier due to a simpler loss landscape, which could allow us to use algorithms such as Bayesian optimization. It would now be interesting to consider different activation functions with more hyperparameters and analyze the effects of tuning them. Furthermore, a systematic run time comparison for more complex dynamical systems would be helpful to assess the advantages and disadvantages of randomized neural network approaches.

\section*{Acknowledgments}

M.T.\ was supported by the EPSRC Centre for Doctoral Training in Mathematical Modeling, Analysis and Computation (MAC-MIGS) funded by the UK Engineering and Physical Sciences Research Council (grant EP/S023291/1), Heriot--Watt University and the University of Edinburgh. We would like to thank D.E.\ Shaw Research for providing the protein molecule data.

\bibliographystyle{unsrturl}
\bibliography{references}

\end{document}